\documentclass[runningheads]{llncs}
\usepackage[T1]{fontenc}
\usepackage[subtle]{savetrees}
\usepackage{graphicx}
\usepackage{orcidlink}
\usepackage{booktabs}
\usepackage{amssymb}
\usepackage{adjustbox}
\usepackage{multirow}
\usepackage{threeparttable}
\usepackage{subcaption}
\usepackage[font=small,labelfont=bf]{caption}

\begin{document}
\title{Time Series Foundation Models \\for Process Model Forecasting}
\titlerunning{Time Series Foundation Models for PMF}
%
\author{Yongbo Yu\orcidlink{0009-0004-2964-6611} 
\and Jari Peeperkorn\orcidlink{0000-0003-4644-4881}  
\and Johannes De Smedt\orcidlink{0000-0003-0389-0275}  
\and Jochen De Weerdt\orcidlink{0000-0001-6151-0504}}
\authorrunning{Y. Yu et al.}
%
\institute{Research Center for Information Systems Engineering (LIRIS), KU Leuven, Belgium
\\ \email{\{FirstName\}.\{LastName\}@kuleuven.be}}
\maketitle              
\begin{abstract}
Process Model Forecasting (PMF) aims to predict how the control-flow structure of a process evolves over time by modeling the temporal dynamics of directly-follows (DF) relations, complementing predictive process monitoring that focuses on single-case prefixes. Prior benchmarks show that machine learning and deep learning models provide only modest gains over statistical baselines, mainly due to the sparsity and heterogeneity of the DF time series. We investigate Time Series Foundation Models (TSFMs), large pre-trained models for generic time series, as an alternative for PMF. Using DF time series derived from real-life event logs, we compare zero-shot use of TSFMs, without additional training, with fine-tuned variants adapted on PMF-specific data. TSFMs generally achieve lower forecasting errors (MAE and RMSE) than traditional and specialized models trained from scratch on the same logs, indicating effective transfer of temporal structure from non-process domains. While fine-tuning can further improve accuracy, the gains are often small and may disappear on smaller or more complex datasets, so zero-shot use remains a strong default. Our study highlights the generalization capability and data efficiency of TSFMs for process-related time series and, to the best of our knowledge, provides the first systematic evaluation of temporal foundation models for PMF.

\end{abstract}
\section{Introduction}
\label{sec:intro}

Business Process Management (BPM) involves designing, executing, monitoring, and improving operational processes. With the increasing availability of event logs and the development of data-driven techniques, Process Mining (PM) has become an essential discipline for monitoring, analyzing, and enhancing real process behavior from execution data. Within PM, Predictive Process Monitoring (PPM) leverages machine learning to predict future process behaviors \cite{ceravolo2024predictive,rama2021deep}, such as the next activity, remaining time, or outcome of an ongoing case. Despite notable progress, PPM mainly focuses on instance-level predictions and therefore offers limited insights into how the overall process structure evolves over time.

Process Model Forecasting (PMF) has been proposed to address this limitation by predicting system-level dynamics \cite{de2023process}, i.e. how the process model itself changes over time. Existing approaches represent process dynamics as time-indexed directly-follows graphs (DFGs) derived from event logs, where each DFG summarizes the control-flow relations observed in a specific time window. Each directly-follows (DF) relation can then be seen as a variable whose frequency evolves over time, so that DF frequencies together form a multivariate time series. Historical DF time series are used to forecast future DF frequencies, which are reassembled into a forecasted DFG that represents the anticipated process model at future time points. Recent work has explored multivariate machine learning and deep learning approaches for PMF~\cite{yu2024multivariate,zhou2025process}, and introduced a unified benchmark pipeline~\cite{yu2025benchmarking} for comparing forecasting methods. These benchmarks show that univariate approaches overall outperform multivariate ones and highlight the particularities of DF time series, including sparsity, heterogeneous seasonal and cyclical effects within the same event log, and patterns that are difficult to capture with a single model configuration.

In parallel, foundation models have transformed learning paradigms across domains. Trained on large and diverse datasets with self-supervised objectives, they provide general-purpose representations that can be adapted to many downstream tasks with limited task-specific training~\cite{bommasani2021opportunities}. In PM, Large Language Models (LLMs) have been applied to, among others, interpret business processes \cite{kourani2025leveraging} and generate suffix predictions \cite{oyamada2025domain,pasquadibisceglie2024lupin}, showing their potential for semantic understanding. However, temporal foundation models remain largely unexplored within this context. Time Series Foundation Models (TSFMs) such as Chronos \cite{ansari2024chronos}, MOIRAI \cite{woo2024unified}, and TimesFM \cite{das2024decoder} extend the foundation model paradigm to temporal data. Trained on vast collections of heterogeneous time series across domains, TSFMs learn generic temporal representations that enable strong zero-shot forecasting, i.e. accurate predictions for unseen datasets without additional training. Recent work has further adapted TSFMs to specialized domains, for example healthcare signals~\cite{gupta2024low} and energy dispatch~\cite{beichter2025decision}, often through parameter-efficient fine-tuning (PEFT) techniques that provide performance gains on out-of-domain data.

Since process model evolution can be represented as structured time series, TSFMs offer a promising alternative to current PMF methods, which often struggle with data sparsity and complex temporal patterns~\cite{yu2025benchmarking}. Foundation models trained on large and diverse corpora may help address these challenges by transferring temporal structure learned in other domains to DF time series. To the best of our knowledge, this paper is the first to investigate TSFMs for PMF and, more generally, for process data.

We study three TSFM families, Chronos~\cite{ansari2025chronos,ansari2024chronos}, MOIRAI~\cite{liu2025moirai,liu2024moirai,woo2024unified}, and TimesFM~\cite{das2024decoder}, comprising eight model variants across zero-shot, PEFT, and full fine-tuning settings. We evaluate these models on DF time series derived from four public event logs and combine time series accuracy metrics with process-aware assessments of the forecasted process models. Our experiments show that TSFMs generally achieve lower forecasting errors than traditional and specialized models trained from scratch on the same logs, while fine-tuning provides only modest and dataset-dependent additional gains. These observations motivate a systematic analysis of model size, model iteration, adaptation strategy, and model family when applying TSFMs to PMF.

The main contributions of this paper are as follows:
\begin{itemize}
    \item Evidence. We conduct a systematic, cross-family evaluation of TSFMs for PMF on DF time series derived from real-life event logs, and show that off-the-shelf models already outperform strong statistical and learning-based baselines in terms of MAE and RMSE.
    \item Adaptation guidance. We compare zero-shot use, LoRA-based PEFT, and full fine-tuning with respect to accuracy, robustness, and data requirements, and identify when fine-tuning yields reliable gains and when it mainly introduces overfitting. 
    \item Process-aware analysis. We combine time-series accuracy metrics with process-aware evaluation of the forecasted models and relate TSFM performance to statistical characteristics of DF time series, yielding insights into which process dynamics particularly benefit from temporal foundation models and how this can inform the design of future PMF and process mining techniques.
\end{itemize}

The remainder of the paper is structured as follows. Section \ref{sec:background} introduces the background on PMF and TSFMs. Section \ref{sec:method} explains how TSFMs are applied and fine-tuned to PMF. Section \ref{sec:experimental_evaluation} describes the experimental setup and presents the results. Section \ref{sec:discussion} discusses the findings and outlines directions for future research. Section \ref{sec:conclusion} concludes the paper.

\section{Background}
\label{sec:background}

This section introduces PMF, reviews methods for time series analysis and forecasting, and summarizes foundation models for time series and their adaptation via fine-tuning.

\subsection{Process Model Forecasting}
\label{subsec:pmf}
Process mining (PM) provides a data-driven perspective on operational processes by analyzing event logs recorded by information systems. An event log $\mathcal{L} = {\sigma_1, \sigma_2, \dots, \sigma_N}$ consists of a collection of cases. Each case $\sigma_i = \langle e_{i, 1}, e_{i, 2}, \dots, e_{i, T_i} \rangle$ is an ordered sequence of events representing the execution history of one process instance, and each event $e_{i, t}$ records the case identifier, executed activity type, timestamp, and some attributes. The ordered sequence of activities in a case is referred to as a trace. Predictive Process Monitoring (PPM) focuses on forecasting the future behavior of ongoing cases to support proactive decision-making \cite{rama2021deep}. Formally, the goal is to learn a function $f_\theta : \Sigma^{*} \rightarrow \mathcal{Y}$ that maps an observed process prefix $\sigma_i^{(k)} = \langle e_{i, 1}, \dots, e_{i, k} \rangle \in \Sigma^{*}$ to a target variable $y_i^{(k)} \in \mathcal{Y}$, such as the next activity, remaining sequence (suffix), completion time, or final outcome. A wide range of machine learning and deep learning approaches have been successfully applied to learn such functions \( f_\theta \), including recurrent neural networks (e.g., LSTMs \cite{tax2017predictive}), Transformer-based models \cite{wuyts2024sutran}, and more recently, LLMs for next activity and suffix predictions \cite{oyamada2025domain,pasquadibisceglie2024lupin}.

PMF extends the predictive focus from individual cases to the global system dynamics \cite{de2023process}. Instead of predicting the continuation of a single trace, PMF aims to forecast how the process model evolves over time. In \cite{de2023process}, univariate forecasting approaches were proposed, while \cite{yu2024multivariate,zhou2025process} investigated interdependencies among directly-follows (DF) relations and applied multivariate techniques to jointly model the corresponding DF time series. \cite{yu2025benchmarking} further introduced a comprehensive benchmarking framework that compares a wide range of forecasting methods and univariate versus multivariate strategies. Their results show that univariate approaches overall outperform multivariate ones, underscoring the inherent complexity of capturing interdependencies across DF relations and the challenges of PMF for conventional forecasting models.

\subsection{Time Series Analysis and Forecasting}
\label{subsec:tsaf}
A diverse set of methodologies has been developed for forecasting \cite{cheng2025comprehensive,kim2025comprehensive}. Early methods were primarily based on statistical models, such as exponential smoothing \cite{gardner1985exponential} and ARIMA \cite{box2015time}, which offer interpretability but rely on strict assumptions about temporal structure. As datasets grew larger and more complex, machine learning (ML) methods emerged as alternatives capable of modeling nonlinear relationships without requiring explicit parametric forms, including random forest \cite{breiman2001random} and XGBoost \cite{chen2016xgboost}. Although not inherently sequential, ML models can incorporate temporal information through features such as lagged variables. 

More recently, deep learning (DL) has emerged as a powerful paradigm for complex forecasting tasks, particularly for high-dimensional, long-range, or highly nonlinear problems. Neural network architectures such as recurrent neural networks (RNNs) \cite{hopfield1982neural} and long short-term memory (LSTM) networks \cite{hochreiter1997long} explicitly model temporal dependencies through recurrent connections. Transformer-based models \cite{vaswani2017attention} further advance the field by using attention mechanisms to learn long-range dependencies efficiently, making them state-of-the-art for many forecasting tasks \cite{liu2024itransformer,nie2023a,zhang2023crossformer}. Recent benchmarking work by \cite{yu2025benchmarking} provides a comprehensive evaluation of these forecasting techniques in the context of PMF.

\subsection{Foundation Models and Fine-Tuning Techniques}
\label{subsec:fm_ft}
Recent advances in artificial intelligence (AI) have been profoundly influenced by foundation models (FMs), which are trained on vast and diverse datasets using large-scale self-supervised objectives and subsequently adapted to a wide range of downstream tasks with limited task-specific training (i.e., fine-tuning) \cite{bommasani2021opportunities}. LLMs constitute a prominent class of FMs~\cite{zhao2023survey}. Models such as GPT-3 \cite{brown2020language} exhibit strong zero-shot and few-shot generation capabilities, and similar ideas have been extended to vision \cite{radford2021learning}, multi-modal learning \cite{hurst2024gpt,team2023gemini}, and business processes, where LLMs support tasks such as process interpretation \cite{kourani2025leveraging,kubrak2024explanatory} and prediction \cite{oyamada2025domain,pasquadibisceglie2024lupin}.

This progress has motivated the development of foundation models for time series forecasting \cite{liang2024foundation,ma2024survey,zhang2024self}. Early efforts adapted LLMs directly to temporal data through prompt engineering \cite{jin2023time} and cross-modal representation learning \cite{chang2023llm4ts,zhou2023one}. In parallel, dedicated time series foundation models (TSFMs) have been proposed, trained on large collections of heterogeneous time series. TimeGPT \cite{garza2023timegpt} pioneered this approach with a large encoder–decoder Transformer trained on diverse public time series to enable zero-shot forecasting. \cite{yu2025benchmarking} evaluated TimeGPT for PMF, though its closed-source nature limited systematic analysis. Chronos \cite{ansari2024chronos} tokenizes time series to fit into T5 architectures \cite{chung2024scaling,raffel2020exploring} and augments training with synthetic data. TimesFM \cite{das2024decoder} uses a decoder-only Transformer trained on datasets such as Google Trends and Wikipedia page views, employing patch-based representations \cite{nie2023a}. MOIRAI \cite{woo2024unified} introduces the LOTSA dataset and trains a masked-encoder patch-based model that supports multivariate forecasting and frequency-level specialization, while MOIRAI-MoE \cite{liu2024moirai} extends this framework with a mixture-of-experts architecture. Benchmarking over large sets of time series from different domains \cite{li2025tsfm} shows that TSFMs generally outperform LLM-based time series models, and that both categories can surpass task-specific models trained from scratch on individual datasets, as reflected in public leaderboards \cite{aksu2024gift,shchur2025fev}. These findings motivate a focus on open-source TSFMs that are pre-trained directly on large-scale time series corpora.

Once pre-trained, TSFMs, similar to other FMs, can be adapted to specific forecasting tasks through fine-tuning, which updates model parameters using labeled task-specific data, while leveraging the knowledge learned during pre-training. Common fine-tuning strategies include full model fine-tuning, where all parameters are updated, and parameter-efficient fine-tuning (PEFT) \cite{han2024parameter} which modifies only a small subset of the model's weights. Two main categories of PEFT methods are selective PEFT, which updates only targeted parameter subsets such as the bias terms or the last few layers, and additive PEFT, which inserts lightweight adapter modules between existing Transformer blocks to achieve task adaptation with minimal architectural changes \cite{zhang2025parameter}. LoRA \cite{hu2022lora} introduces a reparameterization mechanism by inserting trainable low-rank decomposition matrices into selected weight matrices in the self-attention modules. In the context of TSFMs, \cite{gupta2024low} applies LoRA to healthcare time series, and \cite{gupta2024beyond} extends this work by exploring more PEFT methods. \cite{beichter2025decision} further combines decision-focused learning with LoRA to enhance TSFM performance for dispatch tasks. These studies demonstrate the effectiveness of PEFT in improving TSFM performance on out-of-domain data. Since PMF, like many forecasting tasks, involves relatively small domain-specific datasets, the low-rank update structure could potentially enable the model to adapt to process dynamics while mitigating overfitting and preserving the inference efficiency of the original pre-trained model. 

Motivated by these findings, this work investigates zero-shot use, LoRA-based PEFT, and full fine-tuning of TSFMs on DF time series for PMF.

\section{Process Model Forecasting using Time Series Foundation Models}
\label{sec:method}
In this section, we explain how we represent process model evolution as time series derived from event logs, describe the time series foundation models (TSFMs) used in our study, and detail the zero-shot and fine-tuning settings considered.

\subsection{From Event Logs to Process Model Forecasts}
A time series is an ordered sequence of observations recorded at regular time intervals, denoted as $\{y_t\}_{t=1}^T$, where $t=1, 2, \dots, T$. Each data point reflects the state at a specific moment, capturing how a phenomenon evolves over time. Forecasting aims to predict future values based on historical observations. The $h$-step-ahead forecast can be expressed as $\tilde{y}_{t+h} = f(y_t, y_{t-1}, \dots)$, where $h$ is the forecast horizon and $f(\cdot)$ represents a statistical or machine learning model that captures temporal dependencies and patterns in the series. The goal of time series forecasting is to support decision-making by providing insight into uncertain future outcomes.

Given an event log $\mathcal{L}$ as defined in Section~\ref{subsec:pmf}, we derive directly-follows (DF) relations between activities and model their temporal evolution. For two activity labels $a_i$ and $a_j$, we denote by $>_{\mathcal{L}}(a_i,a_j) \in \mathbb{N}$ the number of times $a_i$ is immediately followed by $a_j$ across all traces in $\mathcal{L}$. The corresponding directly-follows graph (DFG) is $DFG_{\mathcal{L}} = (V, E), ~ V = \{ a_i \in \mathcal{A} \}, ~ E = \{ (a_i, a_j, w_{ij}) \mid w_{ij} = >_{\mathcal{L}}(a_i, a_j) \}$, where nodes $V$ represent activity labels and edge weights $w_{ij}$ indicate DF frequencies. 

To model the evolution of these relations over time, the event log $\mathcal{L}$ can be partitioned into a sequence of sublogs over (equal) time intervals $\Delta T$, $\{ \mathcal{L}_{t_1}, \mathcal{L}_{t_2}, \dots, \mathcal{L}_{t_T} \}$, from which corresponding time-indexed DFGs $\{ DFG_{\mathcal{L}_{t_1}}, DFG_{\mathcal{L}_{t_2}}, \dots, DFG_{\mathcal{L}_{t_T}} \}$ are derived. Each DFG represents the process structure observed during the respective time window. The aggregated DF relations can therefore be interpreted as multivariate time series. Formally, the objective of PMF is to learn $f_\theta : \{ DFG_{t_i} \mid i = 1, \dots, T \} \rightarrow \{ DFG_{t_i} \mid i = T+1, \dots, T+n \}$ predicting future process models from their historical evolution. Equivalently, the problem can be reformulated at the level of DF time series, treating each DF relation as a univariate series: $f_\theta : \{ >_{\mathcal{L}_{t_i}}(a_p, a_q) \mid i = 1, \dots, T \} \rightarrow \{ >_{\mathcal{L}_{t_i}}(a_p, a_q) \mid i = T+1, \dots, T+n \}$, where each $>_{\mathcal{L}_{t_i}}(a_p, a_q)$ represents a time-dependent edge weight capturing the temporal dynamics of activity transitions.

\subsection{Time Series Foundation Models}
\label{subsec:tsfm}
Time series foundation models have demonstrated substantial performance gains over traditional machine learning and deep learning models on diverse forecasting tasks, as they can adapt to many different time series profiles and support flexible context windows and prediction horizons. To assess their applicability to DF time series, we consider three representative TSFM families that cover major architectural trends: Chronos, MOIRAI, and TimesFM, spanning encoder–decoder, encoder-only, and decoder-only designs.

\textbf{Chronos} models tokenize time series through scaling and quantization and train Transformer architectures using language-modeling objectives in the first generation, Chronos-T5 \cite{ansari2024chronos}. It employs a T5-style encoder–decoder architecture \cite{raffel2020exploring} and produces probabilistic forecasts by sampling future trajectories. Chronos-Bolt extends this approach with input patching \cite{nie2023a}, which divides historical sequences into non-overlapping chunks to preserve local information and reduce the computational complexity of the attention mechanisms. The decoder then directly generates quantile forecasts across multiple future steps. Recently, Chronos-2 \cite{ansari2025chronos} unifies univariate, multivariate, and covariate-informed forecasting within a single encoder-only architecture, leveraging group attention for efficient in-context cross-learning \cite{das2024context,dong2024survey} across related series and covariates.

\textbf{MOIRAI} models, including 1.0 and 1.1 \cite{woo2024unified}, flatten multivariate inputs using masked encoders with any-variate attention and use multi-patch projection layers to flexibly accommodate different temporal resolutions. It models outputs with mixture distributions, supporting a broad range of downstream tasks. To address limitations in frequency specialization, MOIRAI-MoE \cite{liu2024moirai} incorporates sparse mixture-of-experts routing for fine-grained token-level specialization. MOIRAI-2.0 \cite{liu2025moirai} adopts a decoder-only design with a quantile loss objective and multi-token generation.

\textbf{TimesFM} \cite{das2024decoder} is a decoder-only TSFM that also leverages input patching to efficiently handle long histories. TimesFM 1.0 focuses on point forecasts, TimesFM 2.0 extends the context length, and the latest TimesFM 2.5 further increases scalability while enabling continuous quantile prediction with fewer parameters.

\begin{table}[htbp]
\centering
\caption{Overview of selected TSFM families.}
\label{tab:tsfm_overview}
\begin{adjustbox}{width=\textwidth}
\begin{tabular}{c|c|c|l|l|c|r}
\toprule
\textbf{Model}                    & \multicolumn{1}{c|}{\textbf{Variant}} & \multicolumn{1}{c|}{\textbf{Sizes (Parameters)}} & \multicolumn{1}{c|}{\textbf{Architecture}} & \multicolumn{1}{c|}{\textbf{Forecast Type}} & \multicolumn{1}{c|}{\textbf{Training Data (Observations)}} & \multicolumn{1}{c}{\textbf{Date of Release}}\\ 
\midrule
\multirow{3}{*}{Chronos} & T5                                     & 8M, 20M, 46M, 200M, 710M                & encoder-decoder                   & autoregressive sampling            & 84B   & 13-Mar-24                                            \\ 
                         & Bolt                                   & 9M, 21M, 48M, 205M                      & encoder-decoder                   & direct quantile                    & 100B  & 26-Nov-24                                            \\ 
                         & 2                                      & 120M                                    & encoder-only                      & direct quantile                    & not explicitly stated (mostly synthetic)
                         & 20-Oct-25\\ 
\midrule
\multirow{3}{*}{MOIRAI}  & 1.0 \& 1.1                             & 14M, 91M, 311M                          & masked-encoder                    & direct quantile                    & 27B                                       & 19-Mar/14-Jun-24        \\ 
                         & MoE                                    & 11M/117M, 86M/935M                      & decoder-only + MoE                & direct quantile                    & 27B                      & 1-Nov-24                         \\ 
                         & 2.0                                    & 11.4M                                   & decoder-only                      & autoregressive quantile            & 296B                      & 8-Aug-25                 \\ 
\midrule
\multirow{3}{*}{TimesFM} & 1.0                                    & 200M                                    & decoder-only                      & autoregressive point               & 100B         & 2-Feb-24                        \\ 
                         & 2.0                                    & 500M                                    & decoder-only                      & autoregressive point               & \textgreater{}100B (not explicitly stated)      & 30-Dec-24  \\
                         & 2.5                                    & 200M                                    & decoder-only                      & autoregressive quantile            & \textgreater{}100B (not explicitly stated)      & 15-Sep-25  \\ 
\bottomrule
\end{tabular}
\end{adjustbox}
\end{table}

Table~\ref{tab:tsfm_overview} summarizes the selected TSFM variants, their sizes, architectures, forecast types, training data, and release dates. For MOIRAI-MoE, the model sizes (e.g., 86M/935M) indicate the number of activated versus total parameters. Across these families, a clear trend is the shift toward quantile forecasting and training on increasingly large and heterogeneous pre-training datasets. Notably, recent models achieve improved performance with fewer parameters, as reported in \cite{aksu2024gift,ansari2025chronos,liu2025moirai}.

\subsection{Zero-Shot and Fine-Tuning for TSFMs}
\label{subsec:zeroshot_finetuning}
To understand how TSFMs adapt to DF time series, we evaluate three commonly used settings: zero-shot forecasting, LoRA-based parameter-efficient fine-tuning (PEFT), and full fine-tuning. 

In the zero-shot setting, pre-trained TSFMs are applied directly to DF time series without any additional training. This setting evaluates how well pre-trained temporal representations transfer to data that differ substantially from the natural, economic, and sensor series typically found in TSFM training corpora. Since DF time series encode process behavior rather than physical or financial dynamics, and our DF time series exhibit characteristics that differ substantially from commonly used public time series datasets (see Section \ref{subsec:analysis} for detailed analysis), zero-shot performance provides insight into the robustness of these models to domain shifts.

LoRA (low-rank adaption) \cite{hu2022lora} introduces lightweight, trainable low-rank matrices that reparameterize weight updates inside Transformer layers while keeping the original weights frozen. Concretely, for a weight matrix $W_{0} \in \mathbb{R}^{d \times k}$, LoRA represents the update as $\Delta W = B A$, where  $B \in \mathbb{R}^{d \times r}$ and $A \in \mathbb{R}^{r \times k}$ are trainable matrices of rank $r \ll \min(d,k)$, typically initialized such that $A_0 \sim \mathcal{N}(0,1)$ and $B_0 = 0$. During training, the original weight $W_{0}$ remains frozen and only the low-rank factors $A$ and $B$ are optimized. Hence, the effective weight used by the model becomes $W = W_{0} + \frac{\alpha}{r} B A$, where $\alpha$ is a scaling factor that stabilizes training when the rank $r$ is small. In Transformer architectures, LoRA is typically applied to the query (Q), key (K), value (V), and output (O) projection matrices of the self-attention mechanism \cite{dettmers2023qlora,hu2022lora}, though it can also be used in feedforward layers \cite{biderman2024lora}. Because the number of trainable parameters is proportional to $r(d+k)$ rather than $dk$, LoRA significantly reduces memory usage and training cost. This approach aims at preserving the general knowledge encoded in the frozen pre-trained weights while allowing the adapters to capture task-specific temporal patterns related to DF time series.

Full fine-tuning updates all model parameters using the DF time series. While this approach is computationally more demanding than parameter-efficient methods, it remains feasible given the moderate size of both our datasets and the selected TSFMs (especially when contrasted with large language models). However, the limited amount of task-specific training data increases the risk of overfitting. In principle, full fine-tuning allows the model to fully specialize to DF time series dynamics and can be viewed as an upper bound on achievable task-specific performance.

\section{Experimental Evaluation}
\label{sec:experimental_evaluation}
In this section, we evaluate TSFMs for the PMF task. 
First, we discuss the experimental setup and models used, then we present an initial time series analysis of DF characteristics, and finally we report the predictive results.

\subsection{Experimental Setup}
\label{subsec:experimental_setup}

\subsubsection{Data}
To evaluate the TSFMs for PMF, we select four publicly available event logs: BPI Challenge 2017~\cite{bpic2017}, BPI Challenge 2019~\cite{bpic2019}, the Sepsis event log~\cite{sepsis}, and a hospital billing event log~\cite{hospitalbilling}. The BPI2019 event log includes four flow types. In this study, we use the sublog corresponding to the ``3-way match, invoice before GR'' (2018-01-01 to 2019-01-27), denoted as BPI2019\_1. Preprocessing, transformation, and out-of-time splitting follow \cite{yu2025benchmarking}. Table \ref{tab:stats_logs} summarizes the statistics of the four processed event logs. We aggregate by day (timestep = 1 day) and forecast a 7-day horizon. The preprocessed dataset used in this study is publicly available in \cite{yu_2026_18327515}. 

\begin{table}[htbp]
  \centering
  \caption{Summary statistics of the processed event logs.}
  \label{tab:stats_logs}
  \begin{adjustbox}{width=\textwidth}
    \begin{tabular}{l|c|r|c|r|c|c}
    \toprule
    \multicolumn{1}{c|}{\textbf{Event logs}} & \textbf{Variants} & \multicolumn{1}{c|}{\textbf{Cases}} & \textbf{Activities} & \multicolumn{1}{c|}{\textbf{Events}} & \textbf{DFs (variables)} & \textbf{Time length (days)} \\ 
        \midrule
BPI2017 & 29 & 40,229 & 10 & 248,236 & 21 & 319 \\
BPI2019\_1 & 741 & 197,521 & 32 & 1,298,887 & 149 & 307 \\
Sepsis & 790 & 999 & 18 & 16,009 & 135 & 459 \\
Hospital Billing & 314 & 78,828 & 17  & 570,803 & 73 & 726 \\
    \bottomrule
    \end{tabular}
    \end{adjustbox}
\end{table}

Following \cite{li2025tsfm}, longer look-back windows can improve TSFM performance; therefore, during inference, we use an expanding window (all prior data up to the current timestep) to maximize historical context. Because Transformer inputs are typically fixed for efficient batching, training uses a sliding context window of 48 days (moving one step per sample). For all evaluations, we use the last 20\% of each series as the test set (windowed, one-step moves). For fine-tuning, this means: 60\% train, 20\% validation, 20\% test.

\subsubsection{Model and Fine-Tuning Selection}
We structure our experiments to answer the following questions:
\begin{enumerate}
	\item \textbf{Model size}: Do larger foundational model sizes generally yield better performance (within the same family)?
	\item \textbf{Model iteration}: Do newer foundational models improve performance? 
	\item \textbf{Adaption strategies:} Do LoRA or full fine-tuning provide performance gains over zero-shot inference?
	\item \textbf{Model families:} Is there a model family or variant that significantly outperforms others across datasets?
\end{enumerate}

For zero-shot evaluation, we include four sizes of Chronos-Bolt, two sizes of MOIRAI-1.1, Chronos-2, MOIRAI-MoE-base, MOIRAI-2.0, and TimesFM (1.0, 2.0 and 2.5). For LoRA and full fine-tuning, we focus on two sizes of Chronos-Bolt and two sizes of MOIRAI-1.1, and Chronos-2 as they support full fine-tuning.
The goal is to disentangle the fine-tuning capabilities between small and larger models. Most models are univariate; some (MOIRAI-1.1, MOIRAI-MoE, Chronos-2) support multivariate forecasting. However, to keep the comparison consistent and because univariate results were shown to generally outperform multivariate approaches in PMF \cite{yu2025benchmarking} and many other forecasting tasks \cite{abdelmalak2025channel,han2024capacity,nie2023a}, we use univariate inference for all models\footnote{In initial experiments, multivariate models also did not outperform their univariate counterparts.}.

For the LoRA training, given the moderate size of the selected TSFMs and the overfitting risk in PMF observed in \cite{yu2025benchmarking}, we follow the findings and recommendations of \cite{biderman2024lora,hu2022lora}. We set a small rank $r=2$ with scaling factor $\alpha=4$, and apply LoRA to the four weight matrices $W_q$, $W_k$, $W_v$, $W_o$ in the self-attention module. The patch size is fixed at 16 and the batch size at 32. We use a learning rate of 1e-4 and train for 3 epochs with the AdamW optimizer. All other hyperparameters follow the original model checkpoints. For full fine-tuning, we mainly follow the method and settings in \cite{ansari2024chronos,woo2024unified}, while keeping the patch size of 16 and the batch size of 32 for consistency. For inference with zero-shot, LoRA-adapted, and fully fine-tuned models, some models (Chronos, MOIRAI, and TimesFM-2.5) produce probabilistic quantile forecasts. For these models, we generate 100 sample trajectories for three quantile levels (0.1, 0.5, 0.9), and take the median values as the point predictions when applicable.

\subsubsection{Evaluation Criteria}
We evaluate point-forecast accuracy with MAE and RMSE:
\begin{equation}
\mathrm{MAE}=\frac{1}{|\mathcal{T}||\mathcal{D}|} \sum_{t \in \mathcal{T}} \sum_{d \in \mathcal{D}}\left|y_{t, d}-\hat{y}_{t, d}\right|,
\end{equation}
\begin{equation}
\mathrm{RMSE}=\frac{1}{|\mathcal{D}|}\sum_{d \in \mathcal{D}} \sqrt{\frac{1}{|\mathcal{T}|} \sum_{t \in \mathcal{T}} \left(y_{t, d}-\hat{y}_{t, d}\right)^2}.
\end{equation}

For a process-aware evaluation, we use Entropic Relevance (ER) \cite{alkhammash2022entropic} as adapted in \cite{yu2025benchmarking} to handle incomplete traces. ER is a stochastic process conformance measure which quantifies the average number of bits required to encode traces from the event log. Models that closely reflect observed behaviors require fewer bits (higher relevance), while models that deviate from the log require more bits (lower relevance). ER captures both precision and recall by penalizing both unobserved log variants and model-allowed but unseen behaviors. Consequently, lower ER values indicate process models that encode event logs more concisely and accurately represent process executions.

This experiment follows the framework of \cite{yu2025benchmarking}, allowing the direct comparison with these results as a benchmark. From these results, we select two of the strongest performing baselines: a 7-day lag seasonal naive forecast and a hyperparameter-optimized XGBoost model. All experiments are conducted on a single NVIDIA H100 GPU (80G) using TF32 precision. The full experimental resources are made publicly available in our GitHub repository\footnote{https://github.com/YongboYu/pmf-tsfm}.

\subsection{DF Time Series Analysis}
\label{subsec:analysis} 
To analyze the temporal dynamics of DF time series in a nuanced and comprehensive way, we adopt quantitative measures of seasonality, trend, stationarity, transition, shifting, correlation, and non-Gaussianity, as used in benchmark studies \cite{li2025tsfm,qiu2024tfb,zhang2024probts}, measuring different aspects that inform which time series modeling approach is suitable and whether the data are even useful for time series modeling altogether:
\begin{itemize}
\item Seasonality: recurring patterns at regular intervals.
\item Trend: long-term directional movement.
\item Stationarity: whether statistical properties such as mean and variance are stable over time. 
\item Transitions: abrupt or gradual changes in behavior.
\item Shifting: changes in level or timing, including vertical and horizontal offsets.
\item Correlation: dependence between variables.
\item Non-Gaussianity: departures from normality, such as skewness or kurtosis.
\end{itemize}

The specific formulas for these metrics can be found in \cite{li2025tsfm,qiu2024tfb}. Table \ref{tab:stats_df} reports these characteristics for the DF time series. Compared to the 21 benchmark datasets in \cite{li2025tsfm}, our DF time series show higher transition, shifting, and non-Gaussianity, indicating more complex patterns. Among the datasets, BPI2017 appears most predictable. BPI2019\_1 exhibits very low stationarity and high volatility, shifting, and non-Gaussianity. Sepsis shows low trend and very low stationarity scores, along with high non-Gaussianity, indicating weak signals. Hospital Billing seems relatively predictable with high trend scores, but exhibits high shifting, likely due to its much longer time span compared to the others. Given these vastly different time series characteristics, it is hard to devise a best-in-class model to perform PMF. This further motivates our choice for foundational models, which can recognize and hence produce appropriate forecasts for time series with diverging characteristics, even within one data set because they were trained on a high number of different time series.

\begin{table}[htbp]
  \centering
  \caption{Statistical characteristics of DF time series derived from the processed event logs.}
  \label{tab:stats_df}
  \begin{adjustbox}{width=\textwidth}
    \begin{tabular}{l|c|c|c|c|c|c|c}
    \toprule
    \multicolumn{1}{c|}{\textbf{Dataset}} & \textbf{Seasonality} & \textbf{Trend} & \textbf{Stationarity} & \textbf{Transition} & \textbf{Shifting} & \textbf{Correlation} & \textbf{Non-Gaussianity} \\ 
        \midrule
BPI2017 & 0.734 & 0.255 & 0.222 & 0.059 & 0.269 & 0.646 & 0.334 \\
BPI2019\_1 & 0.698 & 0.154 & 0.094 & 0.145 & 0.362 & 0.690 & 0.465 \\
Sepsis & 0.653 & 0.087 & 0.003 & 0.225 & 0.276 & 0.705 & 0.585 \\
Hospital Billing & 0.603 & 0.260 & 0.137 & 0.171 & 0.483 & 0.620 & 0.457 \\
    \bottomrule
    \end{tabular}
    \end{adjustbox}
\end{table}

\subsection{Predictive Results}
\label{subsec:results}

Tables \ref{tab:results_1_MAE} and \ref{tab:results_1_RMSE} report zero-shot MAE and RMSE (mean $\pm$ standard deviation across DF series), including two baselines. The best baseline is marked with $*$, and percentage changes show mean error relative to that baseline. The best results are in bold, and the second-best are italicized. Table \ref{tab:results_2} reports MAE and RMSE for zero-shot, LoRA, and full fine-tuning on selected TSFMs, following the same formatting conventions. Overall, most TSFMs in the zero-shot setting consistently and significantly outperform the benchmarks, except for some smaller old models on BPI2017. The latest models (Chronos-2, MOIRAI-2.0, and TimesFM-2.5) demonstrate particularly strong improvements. In general, LoRA and full fine-tuning further enhance performance relative to zero-shot results, but gains are dataset-dependent and sometimes inconsistent (occasionally performance degrades).

\begin{table}[htbp]
\centering
\caption{Zero-shot evaluation of TSFMs using MAE.}
\label{tab:results_1_MAE}
\begin{adjustbox}{width=\textwidth}
\begin{tabular}{l|c|c|c|c|c}
\toprule
\multicolumn{1}{c|}{Model} & Size & BPI2017 & BPI2019\_1 & Sepsis & Hospital Billing \\
\midrule
Naive Seasonal & - & 8.30 \(\pm\) 9.72 ($*$) & 14.47 \(\pm\) 68.85 & .117 \(\pm\) .197 ($*$) & 1.77 \(\pm\) 3.23 ($*$) \\
XGBoost & - & 8.50 \(\pm\) 9.56 & 14.70 \(\pm\) 51.83 ($*$) & .169 \(\pm\) .206 & 2.67 \(\pm\) 5.45 \\
\midrule
Chronos-Bolt-tiny & 9M
& 11.64 \(\pm\) 14.06 ($\uparrow$40\%) 
& 13.02 \(\pm\) 49.61 ($\downarrow$10\%) 
& .101 \(\pm\) .194 ($\downarrow$14\%) 
& \textbf{1.39 \(\pm\) 2.65 ($\downarrow$21\%)} \\
Chronos-Bolt-mini & 21M
& 9.70 \(\pm\) 11.55 ($\uparrow$17\%) 
& 12.00 \(\pm\) 48.14 ($\downarrow$17\%) 
& .098 \(\pm\) .187 ($\downarrow$17\%)
& 1.40 \(\pm\) 2.69 ($\downarrow$21\%) \\
Chronos-Bolt-small & 48M
& 7.72 \(\pm\) 8.70 ($\downarrow$7\%) 
& 11.85 \(\pm\) 47.06 ($\downarrow$18\%) 
& .100 \(\pm\) .193 ($\downarrow$15\%) 
& 1.40 \(\pm\) 2.70 ($\downarrow$21\%) \\
Chronos-Bolt-base & 205M
& 7.62 \(\pm\) 8.50 ($\downarrow$8\%) 
& 11.64 \(\pm\) 47.72 ($\downarrow$20\%) 
& .098 \(\pm\) .189 ($\downarrow$16\%) 
& 1.40 \(\pm\) 2.71 ($\downarrow$21\%)\\
Chronos-2 & 120M
& 7.25 \(\pm\) 8.25 ($\downarrow$13\%) 
& 11.39 \(\pm\) 54.21 ($\downarrow$21\%) 
& .090 \(\pm\) .174 ($\downarrow$23\%) 
& \textbf{1.39 \(\pm\) 2.67 ($\downarrow$21\%)} \\
\midrule
MOIRAI-1.1-R-small & 14M
& 10.24 \(\pm\) 12.33 ($\uparrow$23\%) 
& 12.88 \(\pm\) 49.52 ($\downarrow$11\%) 
& .088 \(\pm\) .177 ($\downarrow$25\%)
& 1.48 \(\pm\) 2.79 ($\downarrow$16\%) \\
MOIRAI-1.1-R-large & 311M
& 9.29 \(\pm\) 10.90 ($\uparrow$12\%)
& 12.30 \(\pm\) 48.40 ($\downarrow$15\%)
& .086 \(\pm\) .170 ($\downarrow$26\%)
& 1.43 \(\pm\) 2.71 ($\downarrow$19\%) \\
MOIRAI-MoE-base & 86/935M
& 7.34 \(\pm\) 8.32 ($\downarrow$12\%) 
& 11.37 \(\pm\) 46.74 ($\downarrow$21\%) 
& \textit{.085 \(\pm\) .166 ($\downarrow$27\%)} 
& 1.40 \(\pm\) 2.69 ($\downarrow$21\%) \\
MOIRAI-2.0 & 11.4M
& \textbf{6.87 \(\pm\) 7.66 ($\downarrow$17\%)} 
& \textit{10.99 \(\pm\) 46.56 ($\downarrow$24\%)} 
& \textbf{.084 \(\pm\) .162 ($\downarrow$28\%)} 
& \textbf{1.39 \(\pm\) 2.67 ($\downarrow$21\%)} \\
\midrule
TimesFM-1.0 & 200M
& 7.18 \(\pm\) 7.94 ($\downarrow$14\%) 
& 12.85 \(\pm\) 49.85 ($\downarrow$11\%) 
& .098 \(\pm\) .185 ($\downarrow$16\%) 
& 1.42 \(\pm\) 2.74 ($\downarrow$20\%) \\
TimesFM-2.0 & 500M
& 7.07 \(\pm\) 7.88 ($\downarrow$15\%) 
& 11.54 \(\pm\) 45.79 ($\downarrow$20\%) 
& .097 \(\pm\) .184 ($\downarrow$17\%) 
& 1.42 \(\pm\) 2.71 ($\downarrow$20\%) \\
TimesFM-2.5 & 200M
& \textbf{6.87 \(\pm\) 7.77 ($\downarrow$17\%)} 
& \textbf{10.75 \(\pm\) 44.16 ($\downarrow$26\%)} 
& .096 \(\pm\) .186 ($\downarrow$18\%) 
& 1.42 \(\pm\) 2.74 ($\downarrow$20\%) \\
\bottomrule
\end{tabular}
\end{adjustbox}
\begin{tablenotes}[flushleft]
\scriptsize
\item Values show mean \(\pm\) standard deviation across DF series. Best baseline is marked with $*$, and TSFM percentage changes relative to it are shown in brackets. Best results are bold; second-best italicized.
\end{tablenotes}
\end{table}

\begin{table}[htbp]
\centering
\caption{Zero-shot evaluation of TSFMs using RMSE.}
\label{tab:results_1_RMSE}
\begin{adjustbox}{width=\textwidth}
\begin{tabular}{l|c|c|c|c|c}
\toprule
\multicolumn{1}{c|}{Model} & Size & BPI2017 & BPI2019\_1 & Sepsis & Hospital Billing \\
\midrule
Naive Seasonal & - & 12.43 \(\pm\) 16.84 & 25.58 \(\pm\) 152.31 & .187 \(\pm\) .259 ($*$) & 2.21 \(\pm\) 3.92 ($*$) \\
XGBoost & - & 11.91 \(\pm\) 14.19 ($*$) & 23.87 \(\pm\) 107.90 ($*$) & .209 \(\pm\) .259 & 3.03 \(\pm\) 5.86 \\
\midrule
Chronos-Bolt-tiny & 9M
& 14.41 \(\pm\) 17.33 ($\uparrow$21\%) 
& 21.08 \(\pm\) 108.59 ($\downarrow$12\%) 
& .134 \(\pm\) .218 ($\downarrow$29\%) 
& \textbf{1.70 \(\pm\) 3.16 ($\downarrow$23\%)} \\
Chronos-Bolt-mini & 21M
& 12.27 \(\pm\) 14.42 ($\uparrow$3\%) 
& 19.98 \(\pm\) 107.31 ($\downarrow$16\%) 
& .131 \(\pm\) .212 ($\downarrow$30\%) 
& 1.71 \(\pm\) 3.19 ($\downarrow$23\%) \\
Chronos-Bolt-small & 48M
& 10.27 \(\pm\) 11.98 ($\downarrow$14\%) 
& 19.35 \(\pm\) 100.71 ($\downarrow$19\%) 
& .133 \(\pm\) .218 ($\downarrow$29\%) 
& 1.71 \(\pm\) 3.20 ($\downarrow$22\%) \\
Chronos-Bolt-base & 205M
& 10.08 \(\pm\) 11.71 ($\downarrow$15\%) 
& 18.86 \(\pm\) 98.60 ($\downarrow$21\%) 
& .133 \(\pm\) .216 ($\downarrow$29\%) 
& 1.72 \(\pm\) 3.20 ($\downarrow$22\%) \\
Chronos-2 & 120M
& 9.85 \(\pm\) 11.53 ($\downarrow$17\%) 
& 19.02 \(\pm\) 108.65 ($\downarrow$20\%) 
& \textit{.128 \(\pm\) .205 ($\downarrow$32\%)} 
& \textbf{1.70 \(\pm\) 3.16 ($\downarrow$23\%)} \\
\midrule
MOIRAI-1.1-R-small & 14M
& 13.13 \(\pm\) 15.75 ($\uparrow$10\%) 
& 21.48 \(\pm\) 112.57 ($\downarrow$10\%) 
& .131 \(\pm\) .215 ($\downarrow$30\%) 
& 1.81 \(\pm\) 3.28 ($\downarrow$18\%) \\
MOIRAI-1.1-R-large & 311M
& 11.91 \(\pm\) 13.90 (0\%) 
& 20.79 \(\pm\) 111.76 ($\downarrow$13\%) 
& .129 \(\pm\) .208 ($\downarrow$31\%) 
& 1.75 \(\pm\) 3.21 ($\downarrow$21\%) \\
MOIRAI-MoE-base & 86/935M 
& 9.94 \(\pm\) 11.68 ($\downarrow$17\%) 
& 20.02 \(\pm\) 111.04 ($\downarrow$16\%) 
& \textit{.128 \(\pm\) .205 ($\downarrow$31\%)} 
& 1.72 \(\pm\) 3.18 ($\downarrow$22\%) \\
MOIRAI-2.0 & 11.4M
& \textit{9.39 \(\pm\) 11.02 ($\downarrow$21\%)} 
& \textit{18.58 \(\pm\) 100.69 ($\downarrow$22\%)} 
& \textbf{.125 \(\pm\) .198 ($\downarrow$33\%)} 
& 1.71 \(\pm\) 3.16 ($\downarrow$23\%) \\
\midrule
TimesFM-1.0 & 200M
& 9.66 \(\pm\) 11.19 ($\downarrow$19\%) 
& 20.44 \(\pm\) 102.32 ($\downarrow$14\%) 
& .134 \(\pm\) .211 ($\downarrow$28\%) 
& 1.73 \(\pm\) 3.24 ($\downarrow$22\%) \\
TimesFM-2.0 & 500M
& 9.50 \(\pm\) 11.08 ($\downarrow$20\%) 
& 18.94 \(\pm\) 99.11 ($\downarrow$21\%) 
& .133 \(\pm\) .211 ($\downarrow$29\%) 
& 1.72 \(\pm\) 3.20 ($\downarrow$22\%) \\
TimesFM-2.5 & 200M
& \textbf{9.32 \(\pm\) 11.00 ($\downarrow$22\%)} 
& \textbf{18.12 \(\pm\) 96.97 ($\downarrow$24\%)} 
& .132 \(\pm\) .211 ($\downarrow$30\%) 
& 1.73 \(\pm\) 3.23 ($\downarrow$22\%) \\
\bottomrule
\end{tabular}
\end{adjustbox}
\begin{tablenotes}[flushleft]
\scriptsize
\item Values show mean \(\pm\) standard deviation across DF series. Best baseline is marked with $*$, and TSFM percentage changes relative to it are shown in brackets. Best results are bold; second-best italicized.
\end{tablenotes}
\end{table}

\begin{table}[htbp]
\centering
\caption{MAE and RMSE for zero-shot, LoRA, and full fine-tuning on selected TSFMs.}
\label{tab:results_2}
\begin{adjustbox}{width=\textwidth}
\begin{tabular}{l|c|c|c|c|c|c|c|c|c}
\toprule
\multicolumn{1}{c|}{Model}                &  
Strategy
& \multicolumn{2}{c|}{BPI2017}                        & \multicolumn{2}{c|}{BPI2019\_1}                     & \multicolumn{2}{c|}{Sepsis}                         & \multicolumn{2}{c}{Hospital Billing}               \\
\cline{3-10}
\multicolumn{1}{l|}{}                &           & \multicolumn{1}{c|}{MAE} & \multicolumn{1}{c|}{RMSE} & \multicolumn{1}{c|}{MAE} & \multicolumn{1}{c|}{RMSE} & \multicolumn{1}{c|}{MAE} & \multicolumn{1}{c|}{RMSE} & \multicolumn{1}{c|}{MAE} & \multicolumn{1}{c}{RMSE} \\
\midrule
\multirow{3}{*}{Chronos-Bolt-small} & zero-shot & 7.72 \(\pm\) 8.70             & 10.27 \(\pm\) 11.98            & 11.85 \(\pm\) 47.06           & 19.35 \(\pm\) 100.71           & .100 \(\pm\) .193             & .133 \(\pm\) .218              & 1.40 \(\pm\) 2.70             & 1.71 \(\pm\) 3.20              \\
                                    & LoRA & 7.74 \(\pm\) 8.76             & 10.32 \(\pm\) 12.09            & 12.10 \(\pm\) 47.91           & 19.83 \(\pm\) 105.03           & .105 \(\pm\) .204             & .138 \(\pm\) .229              & \textit{1.38 \(\pm\) 2.62}             & 1.70 \(\pm\) 3.12              \\
                                    & full tune & 8.58 \(\pm\) 9.84             & 11.13 \(\pm\) 12.97            & 11.60 \(\pm\) 47.53           & 19.65 \(\pm\) 108.04           & \textit{.087 \(\pm\) .168}             & \textbf{.127 \(\pm\) .204}              & \textit{1.38 \(\pm\) 2.62}             & 1.70 \(\pm\) 3.13              \\
\midrule
\multirow{3}{*}{Chronos-Bolt-base}  & zero-shot & 7.62 \(\pm\) 8.50             & 10.08 \(\pm\) 11.71            & 11.64 \(\pm\) 47.72           & \textbf{18.86 \(\pm\) 98.60}            & .098 \(\pm\) .189             & .133 \(\pm\) .216              & 1.40 \(\pm\) 2.71             & 1.72 \(\pm\) 3.20              \\
                                    & LoRA & \textit{7.39 \(\pm\) 8.27}             & \textit{9.88 \(\pm\) 11.55}             & 11.55 \(\pm\) 47.06           & 19.36 \(\pm\) 103.81           & .104 \(\pm\) .206             & .138 \(\pm\) .231              & 1.40 \(\pm\) 2.67             & 1.72 \(\pm\) 3.15              \\
                                    & full tune & 7.52 \(\pm\) 8.35             & 10.08 \(\pm\) 11.73            & 11.42 \(\pm\) 47.41           & 19.44 \(\pm\) 107.87           & .088 \(\pm\) .172             & \textbf{.127 \(\pm\) .205}              & \textbf{1.37 \(\pm\) 2.61}             & \textit{1.69 \(\pm\) 3.11}              \\
\midrule
\multirow{2}{*}{Chronos-2}          & zero-shot & \textbf{7.25 \(\pm\) 8.25}             & \textbf{9.85 \(\pm\) 11.53}             & \textbf{11.39 \(\pm\) 54.21}           & \textit{19.02 \(\pm\) 108.65}           & .090 \(\pm\) .174             & \textit{.128 \(\pm\) .205}              & 1.39 \(\pm\) 2.67             & 1.70 \(\pm\) 3.16              \\
                                    & full tune & 7.89 \(\pm\) 9.08             & 10.49 \(\pm\) 12.29            & \textit{11.40 \(\pm\) 46.55}           & 19.09 \(\pm\) 100.67           & .104 \(\pm\) .210             & .140 \(\pm\) .236              & \textbf{1.37 \(\pm\) 2.60}             & \textbf{1.68 \(\pm\) 3.11}              \\
\midrule
\multirow{3}{*}{MOIRAI-1.1-R-small} & zero-shot &  10.24 \(\pm\) 12.33  &  13.13 \(\pm\) 15.75   &  12.88 \(\pm\) 49.52     &  21.48 \(\pm\) 112.57       & .088 \(\pm\) .177  &   .131 \(\pm\) .215    &    1.48 \(\pm\) 2.79     &   1.81 \(\pm\) 3.28     \\
                                    & LoRA & 9.62 \(\pm\) 11.29             & 12.30 \(\pm\) 14.53            & 13.04 \(\pm\) 50.14           & 21.72 \(\pm\) 112.93           & .096 \(\pm\) .209             & .139 \(\pm\) .243              & 1.54 \(\pm\) 2.89             & 1.86 \(\pm\) 3.38              \\
                                    & full tune & 10.21 \(\pm\) 12.30           & 13.16 \(\pm\) 15.84            & 12.89 \(\pm\) 49.59           & 21.50 \(\pm\) 112.59           & .088 \(\pm\) .178             & .131 \(\pm\) .216              & 1.48 \(\pm\) 2.79             & 1.81 \(\pm\) 3.28              \\
\midrule
\multirow{3}{*}{MOIRAI-1.1-R-large} & zero-shot & 9.29 \(\pm\) 10.90            & 11.91 \(\pm\) 13.90            & 12.30 \(\pm\) 48.40           & 20.79 \(\pm\) 111.76           &  \textbf{.086 \(\pm\) .170}             & .129 \(\pm\) .208              & 1.43 \(\pm\) 2.71             & 1.75 \(\pm\) 3.21              \\
                                    & LoRA & 8.27 \(\pm\) 9.11             & 10.84 \(\pm\) 12.28            & 12.27 \(\pm\) 50.32           & 19.95 \(\pm\) 103.56           & .090 \(\pm\) .180             & .132 \(\pm\) .215              & 1.47 \(\pm\) 2.79             & 1.79 \(\pm\) 3.29              \\
                                    & full tune & 9.25 \(\pm\) 10.85            & 11.90 \(\pm\) 13.91            & 23.06 \(\pm\) 88.20           & 32.29 \(\pm\) 147.01           & .114 \(\pm\) .191             & .159 \(\pm\) .239              &           1.43 \(\pm\) 2.72   &  1.75 \(\pm\) 3.21 \\
\bottomrule
\end{tabular}
\end{adjustbox}
\begin{tablenotes}[flushleft]
\scriptsize
\item Values show mean \(\pm\) standard deviation across DF series. Best results are bold; second-best italicized.
\end{tablenotes}
\end{table}

The following analysis addresses the research questions (RQs) outlined in Section \ref{subsec:experimental_setup}.

RQ1 (Model size): Within the Chronos-Bolt and MOIRAI-1.1 families, larger models with more parameters generally achieve more accurate predictions, consistent with prior findings \cite{ansari2024chronos,liu2025moirai,woo2024unified}. However, 
model size alone does not always guarantee better performance.

RQ2 (Model iteration): Newer models often outperform earlier ones, even with fewer parameters, likely due to architectural improvements and larger/more diverse training data (see Table \ref{tab:tsfm_overview}).

RQ3 (Adaptation strategies): LoRA and full fine-tuning can provide performance gains, although the improvements are often marginal and dataset-dependent. This may be due to the relatively small size and complex patterns of our datasets, which limit the stability and effectiveness of adaptation. Overall, LoRA tends to deliver slightly better results than full fine-tuning, possibly due to its ability to mitigate overfitting.

RQ4 (Model families): No single model consistently outperforms others across all datasets; however, TimesFM performs well overall. Given the rapid progress in the field of TSFMs (and the newer models outperforming older ones), the choice of model family should matter less.

\begin{table}[htbp]
\centering
\caption{Entropic relevance (ER) of forecasted DFGs from three zero-shot TSFMs, baseline models, and DFGs discovered from the test and training sets.}
\label{tab:results_ER}
\begin{adjustbox}{width=\textwidth}
\begin{tabular}{lllll}
\toprule
\multicolumn{1}{c}{Model} & \multicolumn{1}{c}{BPI2017} & \multicolumn{1}{c}{BPI2019\_1} & \multicolumn{1}{c}{Sepsis} & \multicolumn{1}{c}{Hospital Billing} \\
\midrule
Truth & 1.00 \(\pm\) 0.08 (100.0\%) & 2.00 \(\pm\) 0.29 (100.0\%) & 6.27 \(\pm\) 4.88 (100.0\%) & 1.86 \(\pm\) 0.22 (100.0\%) \\
Training & 1.15 \(\pm\) 0.11 (99.4\%) & 3.89 \(\pm\) 0.81 (95.9\%) & 15.75 \(\pm\) 10.67 (77.9\%) & 5.83 \(\pm\) 1.10 (83.0\%) \\
\midrule
Naive Seasonal & 1.06 \(\pm\) 0.12 (99.9\%) & 2.40 \(\pm\) 0.36 (99.3\%) & 21.07 \(\pm\) 8.41 (39.4\%) & 2.17 \(\pm\) 0.34 (99.4\%) \\
XGBoost & 1.01 \(\pm\) 0.08 (100.0\%) & 2.39 \(\pm\) 0.39 (100.0\%) & 15.48 \(\pm\) 8.89 (74.6\%) & 2.12 \(\pm\) 0.35 (99.6\%) \\
\midrule
Chronos-2 & 1.09 \(\pm\) 0.13	(99.7\%) & 2.57 \(\pm\) 0.43 (98.7\%) & 30.50 \(\pm\) 10.04 (13.2\%) & 2.43 \(\pm\) 0.38 (98.3\%) \\
MOIRAI-2.0 & 1.09 \(\pm\) 0.13 (99.7\%) & 2.57 \(\pm\) 0.44 (98.8\%) & 34.21 \(\pm\) 9.23 (4.1\%) & 2.52 \(\pm\) 0.41 (98.0\%) \\
TimesFM-2.5 & 1.10 \(\pm\) 0.14 (99.7\%) & 2.54 \(\pm\) 0.45 (98.9\%) & 27.99 \(\pm\) 11.14 (17.5\%) & 2.39 \(\pm\) 0.40 (98.5\%) \\
\bottomrule
\end{tabular}
\end{adjustbox}
\begin{tablenotes}[flushleft]
\scriptsize
\item Values show mean \(\pm\) standard deviation across windows, with average fitting ratios in brackets.
\end{tablenotes}
\end{table}

Table \ref{tab:results_ER} reports the Entropic Relevance (ER) of forecasted DFGs produced by the three best-performing TSFMs in zero-shot mode (Chronos-2, MOIRAI-2.0, TimesFM-2.5). We include two forecasting baselines and two reference DFGs discovered from the ground truth event logs (Truth) and the full training set (Training). All forecasted DFGs correspond to 7-day prediction windows. Each entry shows the mean ER and its standard deviations across windows, followed by the ratio of fitting traces in brackets. Overall, TSFMs achieve ER values comparable to the Naive Seasonal and XGBoost baselines, except on the Sepsis event log, where all three TSFMs exhibit higher ER and very low fitting ratios. This is likely due to the sparse distribution of cases over many days (Table \ref{tab:stats_logs}) and the weak temporal signal in its DF time series (Table \ref{tab:stats_df}), which complicates 7-day forecasting. Higher ER values indicate that behavior in the log is distributed across a wider range of trace variants, including infrequent paths, which suggests lower regularity and greater control-flow diversity. Conversely, lower ER reflects a more stable and coherent process in which most cases follow a limited number of dominant patterns. For Sepsis, the combination of high ER and low fitting ratio suggests that the process is both behaviorally heterogeneous and difficult to represent faithfully with a single process model. This helps explain why predictive models, including TSFMs, find Sepsis more challenging. Compared with the ER of DFGs discovered from the training set on the Hospital Billing log, the forecasted models show a notable improvement. This may stem from the clear trend, large shifting ratio (Table \ref{tab:stats_df}), and long time span (Table \ref{tab:stats_logs}), and all these need models capable of capturing long-term structural changes.

\section{Discussion}
\label{sec:discussion}

From the experimental results, we observe that time series foundation models (TSFMs) in zero-shot forecasting substantially outperform two selected baselines as well as the broader set of forecasting techniques benchmarked in \cite{yu2025benchmarking} on our DF time series, an out-of-domain modality for the TSFMs’ training corpora. The data analysis in Section \ref{subsec:analysis} highlights that DF time series from different event logs exhibit notably distinct properties and patterns. This helps explain why no single model here, nor in \cite{yu2025benchmarking}, consistently performs best across all four event logs. Despite this heterogeneity, TSFMs generalize well and deliver consistently competitive results across the four logs we evaluated, indicating their potential as broadly applicable forecasters for PMF. 

\begin{figure}[htbp]
    \centering
    \includegraphics[width=0.48\textwidth]{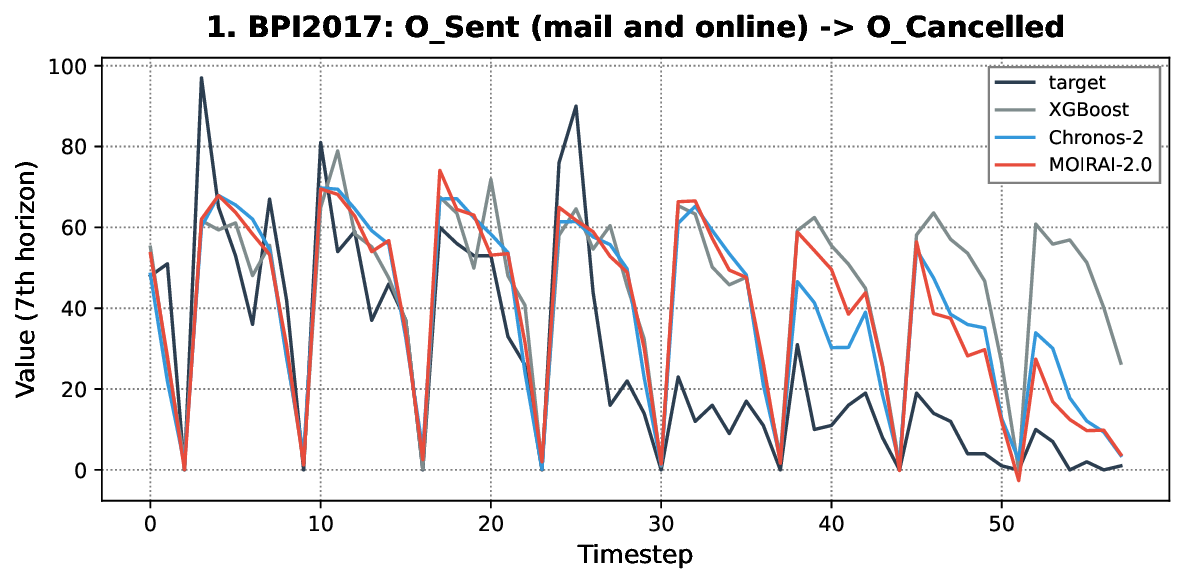}
    \includegraphics[width=0.48\textwidth]{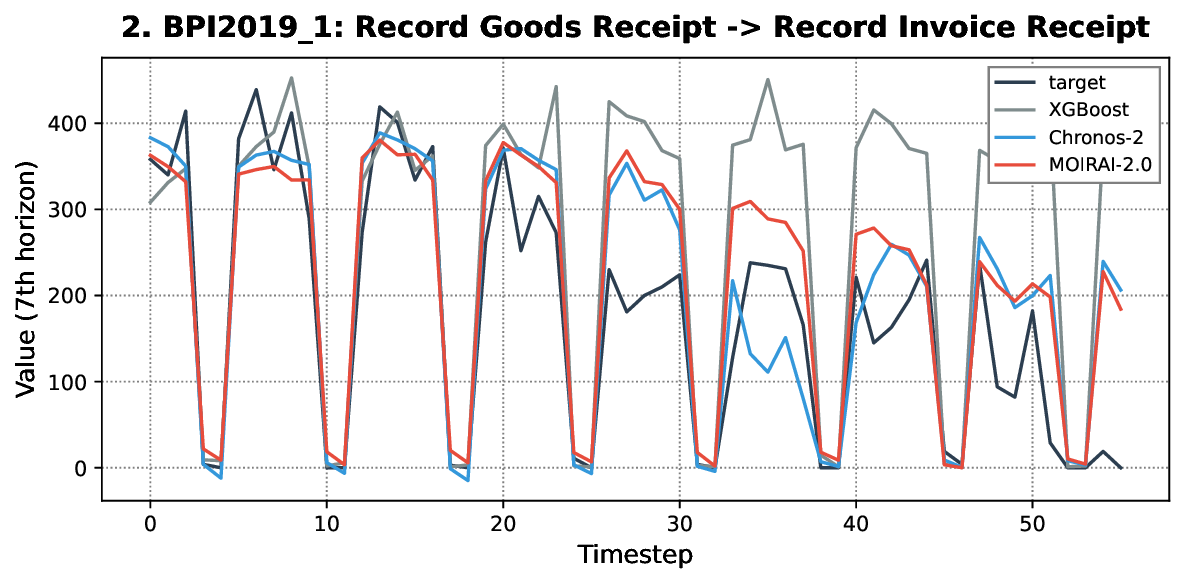}
    \includegraphics[width=0.48\textwidth]{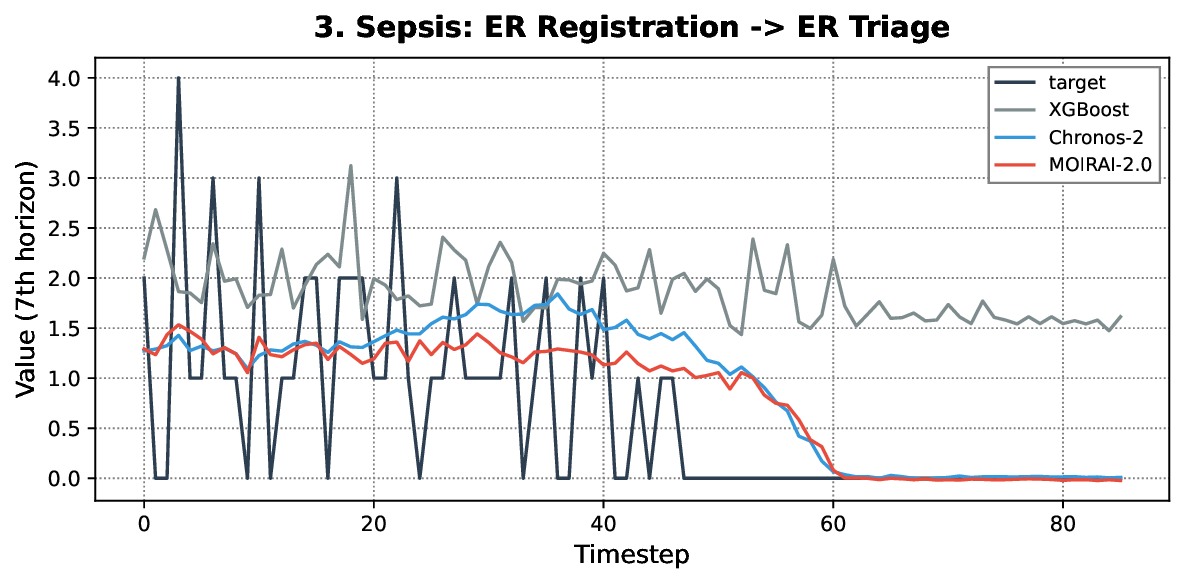}
    \includegraphics[width=0.48\textwidth]{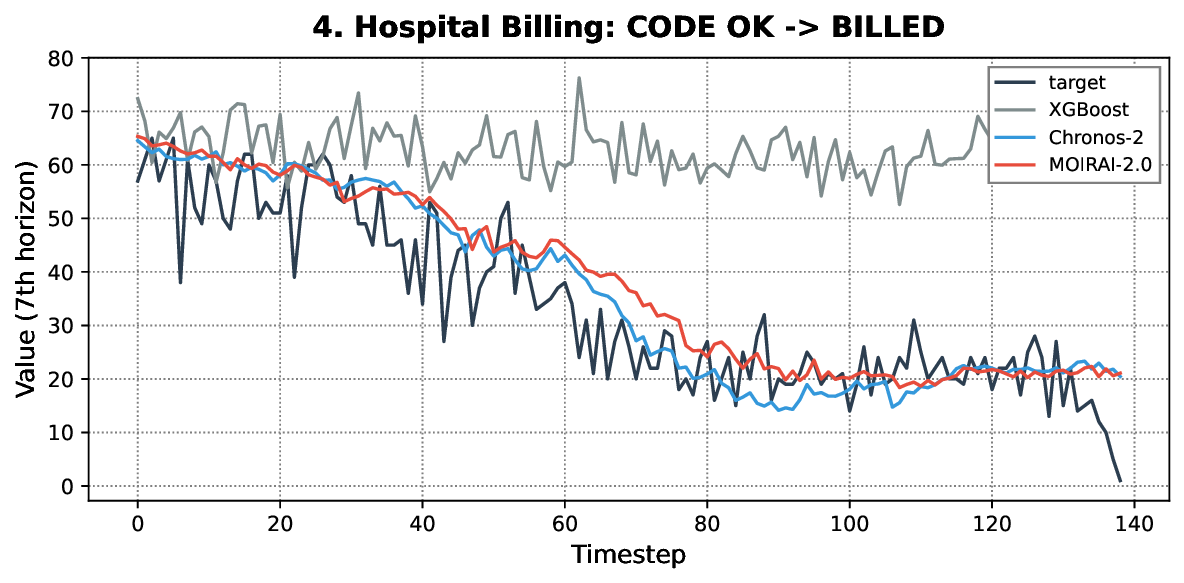}
    \caption{Zero-shot forecasts from Chronos-2 and MOIRAI-2.0 on four DF time series.}
    \label{fig:df_predition_plot}
\end{figure}

Visual inspection of four DF series helps illustrate where TSFMs outperform tree-based ensembles. We plot zero-shot forecasts from Chronos-2 and MOIRAI-2.0 on four DF time series from distinct event logs in Figure \ref{fig:df_predition_plot}. These were selected, due to their particularly large difference between XGBoost and TSFM performance. Forecasts correspond to the next seven days, and we plot the last-day prediction each time against the actual targets. In the first plot, next to the clear seasonal pattern, a sudden decreasing drift occurs before the 30th timestep, posing a challenge for forecasting models. MOIRAI-2.0 and Chronos-2 first miss this drift, but are able to adapt over time, while XGBoost does not. In the second plot, a similar effect is displayed. Again, both TSFMs seem to capture the downward drift better, with Chronos-2 reacting faster and more strongly. In the third plot, DFs in Sepsis are typically infrequent and sparse, and both TSFMs can effectively capture the fading signal in the time series.
In the fourth plot, there is no seasonal effect at play, only a global decreasing effect (drift). Both TSFMs again capture this long-term trend, while the tree-based XGBoost does not.

We also observe that newer, larger TSFMs, and those trained on more diverse corpora, tend to outperform smaller and earlier variants. This aligns with recent findings \cite{edwards2024scaling,shi2024scaling} that show power-law performance gains with increased model and dataset size used for training the foundation. However, scaling parameters alone is beneficial only when sufficient data are available; otherwise, overfitting may degrade performance. This underscores that data scale is often more critical than parameter growth. Regarding adaptation, LoRA and full fine-tuning can improve performance, though gains are not guaranteed, echoing findings in other domains \cite{li2025tsfm}. The effectiveness likely depends on the size and characteristics of the training data. Given the limited size and complex patterns of our datasets, fine-tuning frequently yields marginal improvements and can even hurt performance due to overfitting. Future studies with more (high-quality) event logs may yield further insights.

Some limitations of our study point to natural next steps for future research. We evaluate four event logs; expanding this to a larger, more diverse collection of high-quality logs would strengthen and generalize our findings. Resource constraints and unavailability of source code limited the number of models we fine-tuned and the set of PEFT methods explored. Evaluating additional TSFMs and alternative PEFT approaches could reveal better adaptation strategies for PMF. Moreover, deeper investigation into how architectural choices and pretraining corpora affect performance on process-specific tasks would be valuable. 

Finally, our findings in Figure \ref{fig:df_predition_plot} suggest that concept drift is a key challenge and the drift-score analysis in \cite{yu2025benchmarking} indicate that many event logs are characterized by behavioral shifts over time. Future work should examine how drift detection can be combined with incremental learning or lightweight fine-tuning to support adaptive retraining under non-stationary conditions. In addition, since the current study focuses on directly-follows relations as a control-flow-based representation, a natural next step is to enrich the semantics of execution behaviors, for example by modeling loops and long-range dependencies between activities. Beyond this, an important further step is to extend the approach to additional process dimensions, such as resources, bottlenecks, and decision-related attributes, in order to obtain richer predictive insights.

\section{Conclusion}
\label{sec:conclusion}

In this work, we conducted a comprehensive evaluation of time series foundation models for process model forecasting on directly-follows time series derived from event logs. Our experiments show that TSFMs, even in a zero-shot setting, generally achieve lower forecasting errors than traditional baselines and deliver competitive performance across heterogeneous datasets. Across logs with varying trends, seasonality, stationarity, and shifting patterns, TSFMs generalize well and, as visual inspections of forecasts illustrate, capture long-term dynamics more effectively than conventional models.

We also investigated parameter-efficient fine-tuning via LoRA and full fine-tuning. While both strategies can improve accuracy on some datasets, their effectiveness depends on dataset size and complexity, and on smaller and more complex logs they often yield only modest gains or even degrade performance due to overfitting. In our experiments, architectural innovations and pretraining on larger, more diverse corpora contributed more to performance improvements than model scaling alone.

Overall, our findings indicate that TSFMs are a valuable solution for forecasting the complex temporal behavior of real-world processes and provide a data-efficient default for PMF. This positions temporal foundation models as a practical basis for future process model forecasting approaches and suggests that further work should explore richer structural representations, additional domains, and integration of TSFM-based forecasts into interactive process mining tools.

\begin{credits}
\subsubsection{\ackname} This work was supported in part by the Research Foundation Flanders (FWO) under Project 1294325N as well as grant number G039923N, and Internal Funds KU Leuven under grant number C14/23/031.
\end{credits}
%
%
%
\clearpage
\bibliographystyle{splncs04}
\bibliography{mybibliography}

\end{document}